 \documentclass[sigconf]{aamas} 

\usepackage{balance} 
\usepackage{tikz}
\usepackage{algorithm}
\usepackage{algpseudocode}
\usepackage{colortbl}
\definecolor{taupegray}{rgb}{0.55, 0.52, 0.54}
\usepackage{booktabs}
\usepackage{listings}
\usepackage[title]{appendix}
\usepackage{algorithm}

\setcopyright{ifaamas}
\acmConference[AAMAS '24]{Proc.\@ of the 23rd International Conference
on Autonomous Agents and Multiagent Systems (AAMAS 2024)}{May 6 -- 10, 2024}
{Auckland, New Zealand}{N.~Alechina, V.~Dignum, M.~Dastani, J.S.~Sichman (eds.)}
\copyrightyear{2024}
\acmYear{2024}
\acmDOI{}
\acmPrice{}
\acmISBN{}

\acmSubmissionID{923}

\title{NovelGYM - An Ecosystem for NeuroSymbolic Learning in \\ 
Open World Scenarios}
\title{NovelGym - A Flexible Ecosystem for Hybrid Planning and Learning Systems for Novelty Handling Agents in Open Worlds}
\title{NovelGym: A Flexible Ecosystem for Hybrid Planning and Learning Agents Designed for Open Worlds
}

\author{Shivam Goel}
\affiliation{
  \institution{Tufts University}
  \city{Medford}
  \country{USA}}

\author{Yichen Wei}
\affiliation{
  \institution{Brown University}
  \city{Providence}
  \country{USA}}
\author{Panagiotis Lymperopoulos}
\affiliation{
  \institution{Tufts University}
  \city{Medford}
  \country{USA}}
\author{Klara Chura}
\affiliation{
  \institution{Tufts University}
  \city{Medford}
  \country{USA}}
  
    \author{Matthias Scheutz}
\affiliation{
  \institution{Tufts University}
  \city{Medford}
  \country{USA}}

  \author{Jivko Sinapov}
\affiliation{
  \institution{Tufts University}
  \city{Medford}
  \country{USA}}

\begin{abstract}

As AI agents leave the lab and venture into the real world as autonomous vehicles, delivery robots, and cooking robots, it is increasingly necessary to design and comprehensively evaluate algorithms that tackle the ``open-world''. To this end, we introduce NovelGym\footnote{Project website and codebase source: \href{https://clarech712.github.io/ng-website/}{NovelGYM}}, a flexible and adaptable ecosystem designed to simulate gridworld environments, serving as a robust platform for benchmarking reinforcement learning (RL) and hybrid planning and learning agents in open-world contexts. The modular architecture of NovelGym facilitates rapid creation and modification of task environments, including multi-agent scenarios, with multiple environment transformations, thus providing a dynamic testbed for researchers to develop open-world AI agents.  

\end{abstract}


\newcommand{\BibTeX}{\rm B\kern-.05em{\sc i\kern-.025em b}\kern-.08em\TeX}

\begin{document}


\pagestyle{fancy}
\fancyhead{}


\maketitle 


\section{Introduction}
As AI research ventures beyond ``closed-worlds'' where agents know all task-relevant concepts in advance, the ability to recognize, learn, and adapt to \textit{conceptually} new situations becomes increasingly important. While significant research effort has been invested in creating ``open-world'' systems~\cite{ding2022robot, stern2022model, sarathy2020spotter, boult2022weibull},
comprehensively evaluating them remains a challenge due to 1) the varying and conflicting interpretations of novelty as a concept~\cite{boult2022weibull, muhammad2021novelty, chadwick2023characterizing}, 2) the varying architectural choices made in designing novelty-aware agents~\cite{DBLP:conf/icdl/GoelSSSS22, balloch2023neuro, wang2023describe, khetarpal2022towards} and 3) the unbounded space of possible novelties that an agent may encounter. 
\begin{figure}[t]
    \centering
    \resizebox{0.9\columnwidth}{!}{\includegraphics[width=0.9\textwidth]{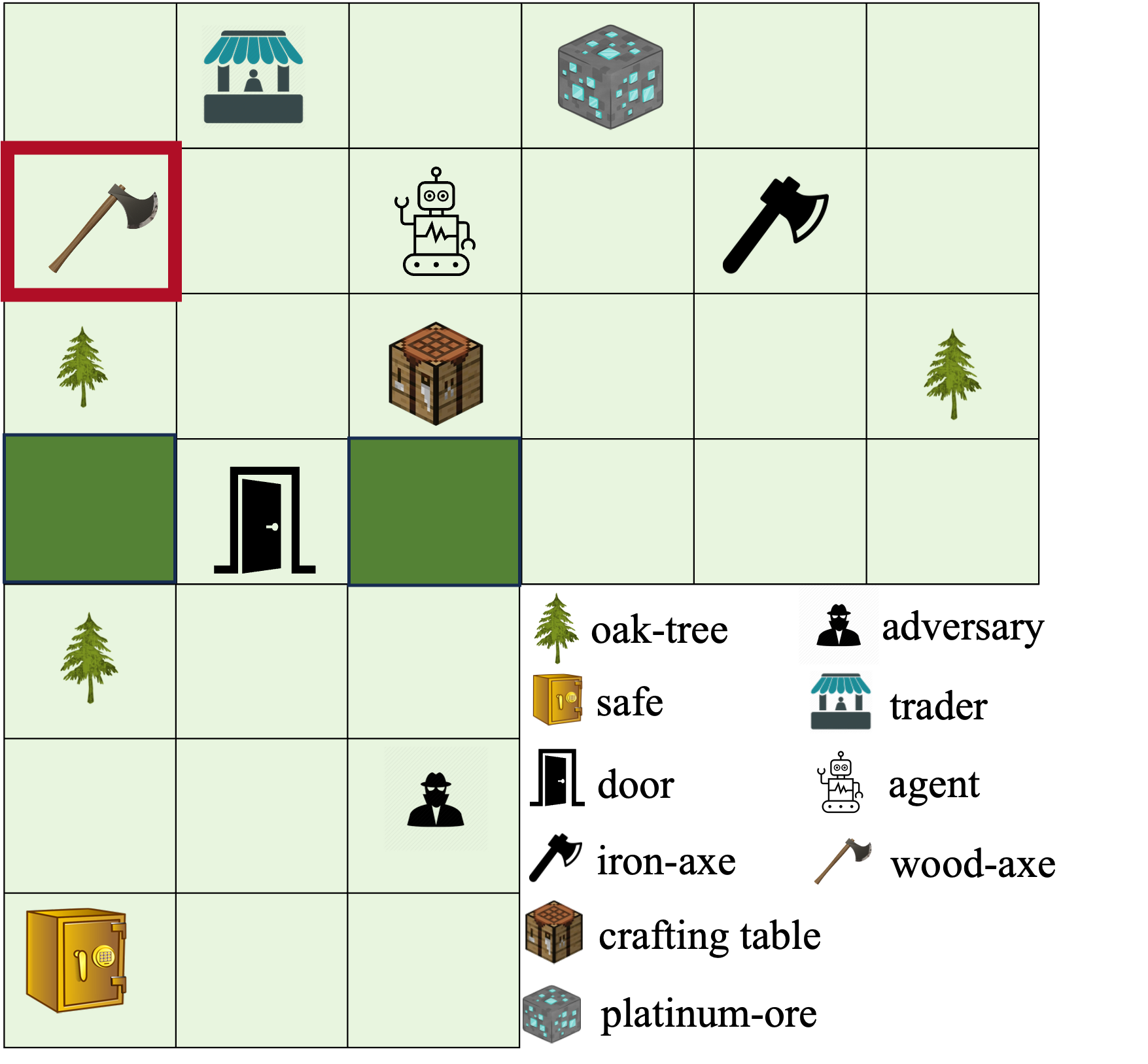}}
    \caption{NovelGym environment representation. The figure shows a gridworld environment with various entities, as described in the legend. The red box highlights the novelty in the environment.}
    \label{fig:novelgym_representation}
\end{figure}

In this work, 
we 
consider {\em novelty} an intrinsically agent-relative concept: An aspect of the world is {\em novel for an agent}, if that agent has not experienced it in the past or cannot derive it from its current knowledge. As such, depending on the particular cognitive and perceptual capabilities of a given agent, different aspects of the world may or may not constitute novelty.  Similarly, a particular aspect of the world that is novel for an agent may also be irrelevant to it, making adaptation unnecessary. For instance, the height of a lamp is a novel concept for an automated vacuum cleaner but one that is irrelevant with respect to its cleaning task. We also emphasize that what may be novel for one agent may not be novel for another. Therefore, when comparing agents' capabilities in novelty adaptation, it is important to control for such differences. As a result, evaluation environments for novelty-aware agents need to be flexible enough to accommodate varying agent architectures, easy to extend to enable rapid development of novelties and tasks, including for multi-agent scenarios, and offer agent-agnostic evaluation metrics that can measure the agent's ability to adapt to novelty compared to non-novelty aware agents of similar capabilities.   

In our work, we propose a new benchmark for the evaluation of novelty-aware agents that is consistent with the aforementioned desiderata. Specifically, NovelGym offers:

\begin{enumerate}
    \item A flexible and modular environment featuring easy task and novelty design for developing and evaluating open-world agents, in single and multi-agent scenarios.
    \item An ecosystem that seamlessly works with agents of different architectures, including symbolic planning agents, reinforcement learners, and hybrid neurosymbolic architectures.
    \item  Benchmarks of various state-of-the-art learning and hybrid methods for novelty handling.
    \item Agent-agnostic evaluation metrics for novelty adaptation.
\end{enumerate}

The rest of our paper is structured as follows: First, we discuss related work on novelty-aware agents and environments to evaluate them, followed by establishing the theoretical framework upon which the design of NovelGym is based. We present the environment architecture and current environment transformations that may serve as novelties for tested agents. We also explain the modular implementation of the environment and how it facilitates further design of tasks, agents, and novelties. Finally, we present evaluation measures and benchmark evaluations of novelty handling agents. 

\section{Related Work}

%
Recently, there has been increasing interest in creating agents that can adapt to sudden and abrupt changes (i.e., novelties).
\citet{klenk2020model} present a trainable model for novelty adaptation called WorldCloner, where a symbolic representation of the pre-novelty world is learned and then used to detect novelties. WorldCloner uses a gridworld where an agent must complete a task that may or may not be obstructed by novelties, but the task is simple and does not involve a complex sequence of operations like crafting or breaking. \citet{stern2022model} also presents a model-based framework named HYDRA that uses a domain-independent planner for the popular video game Angry Birds. \citet{sarathy2020spotter} proposed SPOTTER, an approach that goes beyond pure reinforcement learning methods to learn new operators needed to solve a task when symbolic planning cannot due to novelties. \citet{pardo2020tonic} introduced Tonic, a benchmarking library for deep reinforcement learning that is configurable but limited to the scope of compatible environments and agents. 

Several environments and frameworks for novelties or open-world scenarios have been developed to aid the research and development of those frameworks. NovGrid~\cite{balloch2022novgrid} is a novelty generator built on MiniGrid that allows the injection of novelties into any existing minigrid environment. 
\citet{silver2020pddlgym} present PDDLGym, which is a gym environment for RL research that can be generated from symbolic, PDDL domain files commonly used in planning.
\citet{goel2021novelgridworlds} present NovelGridWorlds, a Minecraft-inspired grid world environment to study the detection and adaptation of novelty that works with Planning and Learning. However, none of the environments provides an easy injection of novelties and the ability to integrate Planning and Learning seamlessly. Our environment provides a modular and highly configurable interface and the flexibility to integrate with both planning and RL agents in multiple ways, which, to the best of our knowledge, doesn't exist. Hence, the contribution of such an ecosystem can enable research in the direction of open-world problem-solving.

\section{Theoretical Framework}
\label{sec:theoretical_framework}
\subsection{Running example}
\label{subsec:running_example}
Let us consider a gridworld as shown in Figure~\ref{fig:novelgym_representation}; the environment is laid out as the two rooms environment separated by a door and two cells (shown in dark green). As shown in Figure~\ref{fig:novelgym_representation}, an agent is facing a \textit{crafting table} (used to craft items such as tree tap, axe, etc.). The agent can collect resources from the environment by moving around with navigation actions (move forward, turn left, turn right). The agent can collect resources by breaking them and then craft them into specific items using certain recipes. For example, an agent can break the oak tree to get two logs, and one log can be crafted into four planks. The agent's goal is to craft a \textit{pogostick}. The recipe for crafting a pogostick involves collecting logs, diamonds, and platinum and crafting various intermediary items such as planks, sticks, tree-tap, etc. There are other entities, such as traders, with whom the agent can interact and \textit{trade} items. The environment also has an adversarial agent that competes with the agent in getting resources to craft the \textit{pogostick}.
\subsection{Environment}
\label{subsec:env}
We formalize the environment $E$ as
$E = \langle G, \mathcal{E}, \mathcal{R}, \mathcal{S}, A, \tau, C \rangle$, where each component is defined as follows:
\paragraph{\textbf{Grid}}
$G \subseteq \mathbb{Z}^2$ represents the set of all grid cells in a 2D gridworld. For an $(m \times n)$ grid, any cell can be uniquely identified by its position $(i, j)$ where $(1 \leq i \leq m)~and~(1 \leq j \leq n)$. For example, the gridworld shown in Figure~\ref{fig:novelgym_representation} consists of $7$ rows and $6$ columns.

\paragraph{\textbf{Entities}}
$\mathcal{E} = \{<t_1, P_{e_1}>, <t_2, P_{e_2}>, \ldots, <t_n, P_{e_n}> \}$ is the set of all the entities in the environment where,
\begin{itemize}
    \item \( t_i \) is the type of the \( i^{th} \) entity from the set \( T \) of all possible entity types: $T = \{ t_1, t_2, \ldots, t_t \}$.
    For instance, an entity ``tree'' might have a type ``oak-tree'' within \( T \).
    
    \item \( P_{e_i} \) is the set of properties of the \( i^{th} \) entity. Each property set is a subset of the global property set $( P_{e_i} \subseteq P )$, which encompasses all possible properties an entity can exhibit: $P = \{ p_1, p_2, \ldots, p_p \}$    
    \item An entity \( e \) is thus represented as a tuple $e = <t, P_e>$ where \( t \in T \) and \( P_e \subseteq P \).
    
    \item Some entities are dynamic, such as other agents or adversaries (refer to Figure 1), which can actively take actions in the environment, influencing state transitions.
    
    \item Entities can be located at individual grid cells, in an agent's inventory, or nested within other entities (e.g., stored inside a chest or safe).
\end{itemize}

For example, as depicted in Figure~\ref{fig:novelgym_representation}, the gridworld environment includes entities like oak-\textit{tree}, representing the tree entity with \textit{oak} type. This tree might have properties such as \textit{breakable}. Similarly, an \textit{axe} entity in the environment might have types like \textit{wood} or \textit{iron} and can possess properties like \textit{graspable}.

\paragraph{\textbf{Recipes}}
$\mathcal{R} = \{r_1, r_2, \dots, r_r\}$ is a set of transformation rules, where a rule $r \in \mathcal{R}$ defines how one or more entities can be transformed into another set of entities.
Each rule is defined as \( r: \mathcal{E}^* \times L_r \rightarrow \mathcal{E}^* \), where $L_r \subseteq G$ represents the locations in the grid world, i.e., a recipe can be applied at different places in the environment. For example, some recipes work only in front of the \textit{crafting table}, or in front of other dynamic entities like \textit{traders} (traders and crafting table shown in Figure~\ref{fig:novelgym_representation}). 
Consider the rule $r: (\{\text{1 platinum 1 stick 2 plank}\}, L_{\text{crafting\_table}}) \to \{\text{1 platinum axe}\}$. It indicates that, 1 unit of \textit{platinum} and \textit{stick}, and two units of \textit{plank} can be crafted into platinum axe at the crafting table. 

\paragraph{\textbf{States}}
$\mathcal{S}$ is the set of all possible states in the environment. For example, as shown in Figure~\ref{fig:novelgym_representation}, a possible state of the environment can be the locations of all the entities in the world.

\paragraph{\textbf{Primitive Actions}}
$A = \{a_1, a_2, \dots, a_a\}$ is the set of all available primitive actions. While all actions are available to the agent in every state, some actions only result in changes to the state in specific contexts. For example, \textit{break} action can be executed anywhere in the gridworld, but would only have an effect if performed in front of an entity that has the \textit{breakable} property (example, a tree).
\paragraph{\textbf{Transition Dynamics}} 
$\tau = \mathcal{S} \times A \rightarrow \mathcal{S}$ denotes the transition dynamics and is responsible for determining the progression from one state to another based on a given action. Formally, the transition dynamics can be represented as:
$$
\tau : \mathcal{S} \times A \times (A_{e_1}, A_{e_2}, \ldots, A_{e_m}) \rightarrow \mathcal{S},
$$
where \( \mathcal{S} \) represents the state space, \( A \) denotes the action space of the primary agent, and the tuple \( (A_{e_1}, A_{e_2}, \ldots, A_{e_m}) \) captures the sequence of actions taken by each dynamic entity in the environment. These dynamic entities can act as adversaries or other environmental actors, influencing the state transitions.

\paragraph{\textbf{Cost function}} 
$ C: \mathcal{S} \times A \rightarrow \mathbb{R}^{+}$, denotes the function that assigns a fixed, non-negative cost to each state and action pair. Specifically, for each action \( a \in A \), in a state $s \in \mathcal{S}$, \( C(s, a) \) provides the associated cost of performing the action in a specific state. For example, the cost associated with moving one step in the grid can be 1, or breaking a tree can have a cost of 5, and depending on the context, it may cost less, for example, breaking when holding a tool.
\begin{figure}
    \centering
    \includegraphics[width=0.47\textwidth]{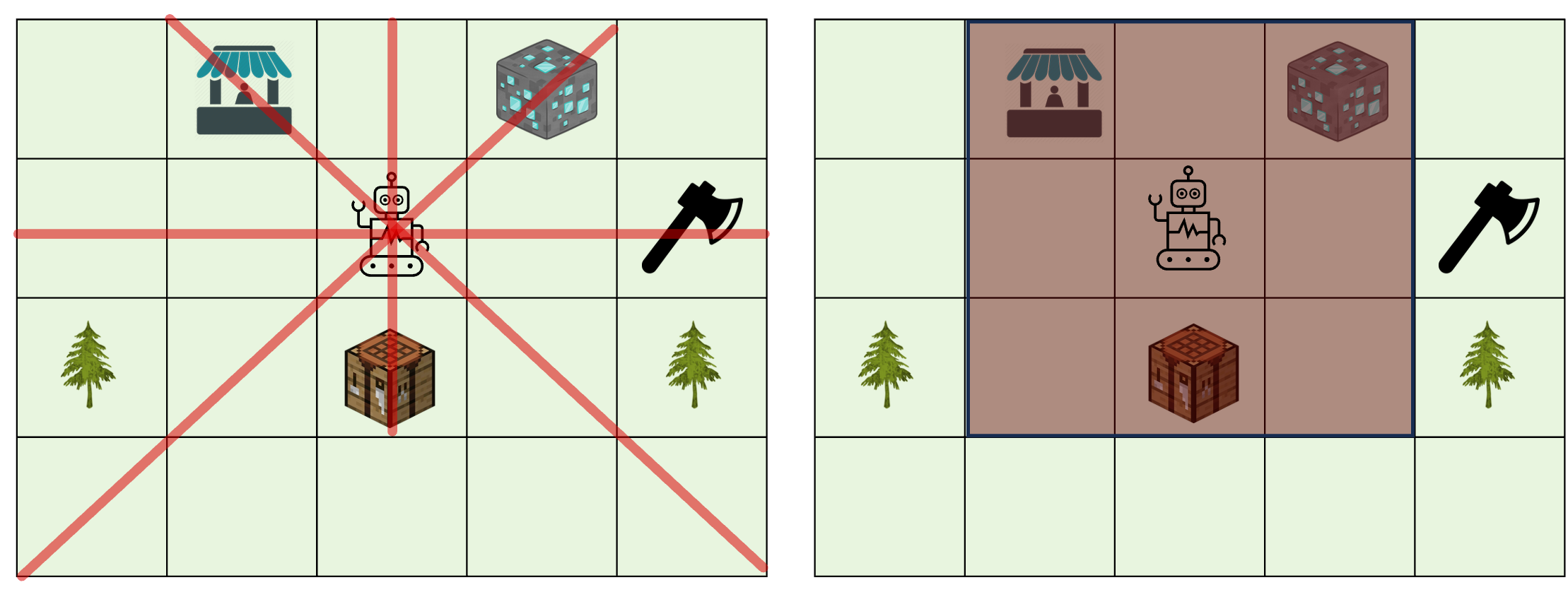}
    \caption{Illustration of the sensor representation of the agent in the environment. (Left) shows a LiDAR representation. (Right) shows an image based local view representation.}
    \label{fig:sensor_representation}
\end{figure}
\begin{table}[t]
\small
\begin{tabular}{@{}ll@{}}
\toprule
\textbf{types}      & \begin{tabular}[c]{@{}l@{}}\texttt{air,crafting\_table,diamond,platinum - physobj}\\ \texttt{tree,diamond,platinum - breakable}\end{tabular} \\ \midrule
\textbf{predicates} & \begin{tabular}[c]{@{}l@{}}\texttt{(holding ?v0 - physobj)}\\ \texttt{(floating ?v0 - physobj)}\\ \texttt{(facing ?v0 - physobj)}\end{tabular}     \\ \midrule
\textbf{fluents}  & \begin{tabular}[c]{@{}l@{}}\texttt{world(?physobj)}\\ \texttt{inventory(?object)}\end{tabular}                                            \\ \midrule
\textbf{action}     & \texttt{approach}                                                                                                                \\ \midrule
\textbf{params}     & \texttt{(?physobj01 ?physobj02)}                                                                                                  \\ \midrule
\textbf{preconditions} & \begin{tabular}[c]{@{}l@{}}\texttt{(and}\\       \texttt{(\textgreater{}= ( world ?physobj02) 1)}\\         \texttt{(facing ?physobj01))}\end{tabular} \\ \midrule
\textbf{effects}    & \begin{tabular}[c]{@{}l@{}}\texttt{(and}\\         \texttt{(facing ?physobj02)}\\         \texttt{(not (facing ?physobj01))}\end{tabular}          \\ \bottomrule
\end{tabular}
\caption{PDDL representation of the symbolic domain.}
\label{tab:symbolic_state}
\end{table}
\subsection{Agent}


We define an agent $\mathcal{A} = \langle Sen, Act, \mathcal{K} ,\Pi \rangle$ as an entity in the environment equipped with sensors $Sen$ to generate observations $O$ of the world, actuators $Act$ that can affect the world, a knowledge repository $\mathcal{K}$ that contains knowledge (learned or provided), and a function $\Pi: O \rightarrow A$ that maps observations to actions. Agent observations can be high-dimensional sub-symbolic vector representations or can be symbolic representations. For example, if the agent has a LiDAR-like sensor (as shown in Figure~\ref{fig:sensor_representation} (left)), then the sensors will produce distances to the objects in the world; similarly, an image-based local view would represent the grid with one hot vector encoding on the entity types (shown in Figure~\ref{fig:sensor_representation} (right)); similarly the mapping can be from a state to a high level symbolic state described using PDDL~\cite{aeronautiques1998pddl}(as shown in Table~\ref{tab:symbolic_state}).

The agent, using actuators $Act$ can act in the environment. The actions can be primitive actions or parameterized actions that implement the primitive-level actions as a sequence. For example, an agent can have an action as \texttt{approach <entity>}, which is implemented by a planner and uses primitive actions (move forward, etc.) to execute the action. The agent's knowledge repository, $\mathcal{K}$, can be initially empty and accumulate knowledge over time. Knowledge can be in the form of parameterized policies or a description of the world symbolically through first-order logic. 
An agent's behavior function $\Pi$ can be implemented based on two popular paradigms for decision-making agents, namely, Planning and Reinforcement Learning, as well as hybrid approaches combining the two.
We further formalize the symbolic planning and reinforcement learning frameworks to further describe different agents.
\subsubsection{Symbolic Planning}
\label{subsec:planning}
In specifying a planning problem, we define $\mathcal{L}$ as a first-order language containing atoms $p(t_1, ..., t_n)$ and their negations $\neg p(t_1, ..., t_n)$, where each atom $t_i$ may be constant or variable.
We define a planning domain in $\mathcal{L}$ as $\mathcal{D} = \langle \tilde{\mathcal{S}}, \mathcal{O}, \tau_\alpha \rangle$, where $\tilde{\mathcal{S}}$ represents the set of symbolic states, $\mathcal{O}$ the set of finite action operators, and $\tau_\alpha$ as the transition function that describes how the state changes as a result of an action operator being executed in the environment \cite{sarathy2018MacGyverACS}. We then define a planning problem as $\mathcal{P} = (\mathcal{D}, \tilde{s}_0, \tilde{S}_g)$, where $\tilde{s}_0$ is the initial state and $\tilde{S_g}$ is the set of goal states. The agent begins in a start state $\tilde{s}_0$ and establishes a plan $\pi$ for reaching one of the goal state contained in $\tilde{S}_g$. 
Hence, the plan $\pi = [o_1, o_2, ..., o_{|\pi|}]$ is a solution to the planning problem $\mathcal{P}$, where each $o_i \in \mathcal{O}$ is an action operator that has a set of preconditions and effects. The preconditions describe the states before executing the operator in the environment, and the effects describe the state of the environment after the agent has executed the operator.
\subsubsection{Reinforcement Learning}
We formalize an RL problem as a Markov Decision Process (MDP) $\mathcal{M} = \langle \mathcal{S}_\beta, A_\beta, \tau_\beta, R, \gamma \rangle$. At time-step $t$, the agent is given a state representation $S_t \in \mathcal{S}_\beta$. Exploring the environment by taking action $A_t \in A_\beta$, the agent is assigned a reward $R_{t+1} \in \mathcal{R} \subset \mathbb{R}$ based on the state $S_{t+1}$ it lands in by choosing an action $A_t$ in a state $S_t$. The goal of the agent is to learn a policy that maximizes the expected return value $G_t = \sum_{k=0}^\infty \gamma^k R_{t+k+1}$ for every state at time $t$. The importance of immediate and future rewards is determined by the discount factor $\gamma \in [0, 1)$.

\subsection{Novelty}
We simulate the open-world using a base environment and a set of environment transformations that act on one (or more) of its constituent elements. The transformations may introduce \textit{novelties} for some agents, depending on their knowledge, perception, and representations. More formally, we can define the  transformation of the environment as a function  $\nu: E \rightarrow E'$. The function $\nu$ may transform the environment as:

\begin{itemize}
    \item \textbf{Layout changes:} The transformation can affect the layout or the grid size $G$ such that $ G \neq G' \text{in}~ E'$. In other words, in the transformed environment $E'$, the layout of the grid $G'$ is not the same as the original layout $G$. 
    \item \textbf{Entity alterations:} New or existing entities may be introduced or modified. This can be represented as a transformation in the set of entities $\mathcal{E}$ such that $\mathcal{E} \neq \mathcal{E}' \text{in}~  E' $.
    \item \textbf{Recipe modifications:} The set of transformation rules \( \mathcal{R} \) can undergo changes, leading to \( \mathcal{R} \neq \mathcal{R}' \) in \( E' \).
    \item \textbf{Action alterations:} The set of available actions \( A \) can experience modifications, resulting in \( A \neq A' \) in \( E' \).
    \item \textbf{Transition dynamics change:} If for some state \( s \in \mathcal{S} \) and \( a \in A \), we have \( \tau(s, a) \neq \tau'(s, a) \) where \( \tau \) and \( \tau' \) are the transition dynamics of \( E \) and \( E' \) respectively, then the transition dynamics have changed.
    \item \textbf{Cost function alterations:} The cost function \( C \) can be modified, leading to $ C \neq C' ~\text{in}~ E'$. This implies differences in costs for certain actions in specific states between the original and transformed environments.

\end{itemize}
Importantly, these transformations are composable, enabling the creation of arbitrarily complex environments. By taking into account an agent's knowledge, perceptions and representations, we can apply transformations and their compositions to create novelties for that agent. For instance, an agent that assumes a specific environment layout for navigation would encounter novelty under a transformation that flips the environment layout, whereas an agent that operates with LiDAR sensors may not. 

%
\begin{figure*}[t]
    \centering
    \includegraphics[width=0.95\textwidth]{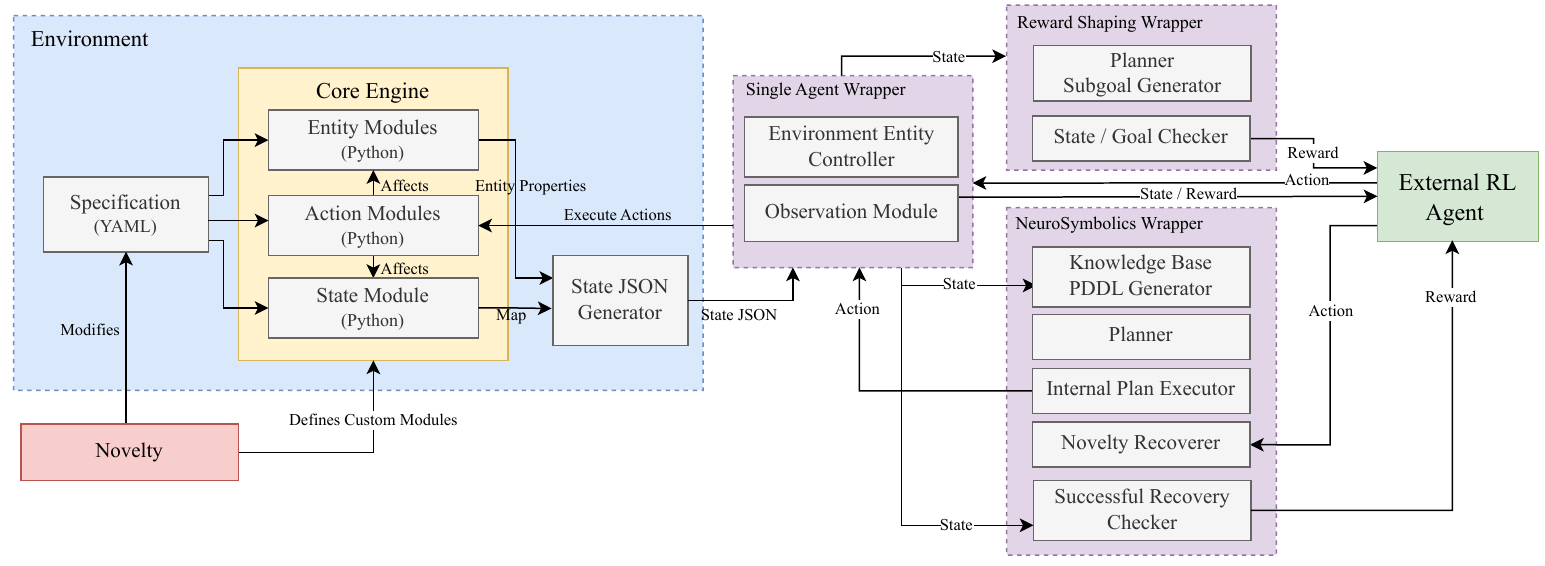}
    \caption{System design of the NovelGym ecosystem. Blue highlights the environment modules, and purple highlights the agent modules.}
    \label{fig:env_interface}
\end{figure*}
\section{NovelGym: Architecture \& Implementation}
The NovelGym ecosystem serves as a platform for developing and evaluating AI agents, with a focus on open-world novelty-aware agents. Comprising a game engine and several aiding modules, the system diagram of the ecosystem can be visualized as shown in Figure~\ref{fig:env_interface}. 
As seen in the figure, NovelGym has two main components. The first component, environment (highlighted in blue), represents the modules that implement the environment and the game engine. The second component, agent (highlighted in purple), showcases the modules that can be used to implement agent architectures. A novelty injector (shown in red) transforms the environment based on the given specifications.
We now describe each component of the system while laying out the important features that enable open-world novelty-aware agent training and evaluations. 
\subsection{Environment}
Our environment implementation is based on PettingZoo~\cite{terry2021pettingzoo} and OpenAI Gym~\cite{brockman2016openai}. The environment component of NovelGym houses a core engine that is responsible for the implementation of entities (entity module shown in Figure~\ref{fig:env_interface}) and actions (action module shown in Figure~\ref{fig:env_interface}). The design of separate modules for all the features of the environment ensures easy task creation and novelty implementation.
The core engine maintains the state of the world through the state module. The configuration of the world can be specified using the specification module. 
Our environment implementation supports multi-agent systems. Thus, each movable entity (agent) takes turns in executing actions. The world map keeps track of the location of entities in the grid world and the coordinates of the rooms. We now describe each module.

\subsubsection{Specification module}
The specification module reads the YAML configuration file and initializes the world. It also loads the action and entity modules and initializes them with the parameters and properties as specified in the configuration file.

\subsubsection{Entity module}
The entity module is responsible for the entities in the environment. Based on the specification, the entities are initialized and may be modified by writing custom entity modules\footnote{A detailed tutorial on how to use and customize the environment can be found \href{https://clarech712.github.io/ng-website/}{here}.}. Entity properties can also be specified directly in a configuration YAML file. For example, objects may be configured to be breakable by hand, by tools, or non-breakable. 


\subsubsection{Action module}
The action module is responsible for the action implementations in the environment. Our environment supports primitive actions as well as higher-level action operators (as described in Section~\ref{subsec:planning}). The action module executes the actions in the environment and ensures that the state of the world is updated by updating the environment.




\subsubsection{World map/state module}
The world map/state module generates the representation of gridworld given the current state. It has a state tracker implemented inside it that keeps track of the entities (including agents). The state tracker also runs scheduled tasks at the end of each timestep. These scheduled tasks may include updating the state for durative actions. For example, the action of break-tree introduces one sapling in the world after three timesteps. In this case, the state tracker will ensure that the world is updated with a sapling after three timesteps of action execution.

\subsubsection{Novelty module}
The novelties are implemented as the extensions or modifications of the environment. The novelty module helps in modifying the entity, action, and state modules depending on the novelty configuration.

\subsection{Agent training architecture}
The agent training architecture is composed of three main components, designed to aid the development of hybrid learning agent architectures. 


\subsubsection{Single agent wrapper}
While our environment supports multi-agent systems, we focus on a single primary agent that solves the task. Other agents in the environment are movable entities that can take actions in the environment. The single-agent wrapper converts the PettingZoo~\cite{terry2021pettingzoo} multi-agent environment into a single-agent one by taking over the action execution of all other agents while exposing the control to the primary agent. The single agent wrapper includes an observation conversion module that converts the state of the world into a desired representation, e.g., a local view of the map, or a lidar representation of the world (see Figure~\ref{fig:sensor_representation}). The customization allows the testing of different state representations. For object-centric observation spaces, the conversion module can be configured to automatically expand the observation and action spaces in cases of additional entities and actions in novelties.

\subsubsection{Neurosymbolic wrapper}
The neurosymbolic wrapper is responsible for combining the symbolic planning agents and reinforcement learning agents. The wrapper maintains a knowledge base that can be in the form of PDDL. With a pre-defined PDDL template, the wrapper automatically generates PDDL files by referring to the information from the pre-novelty configuration file and individual object and action modules. The generated PDDL can be sent to a planner (implemented in the planner component through MetricFF~\cite{hoffmann2003metric}) to generate plans. The plan executor ensures that each operator in the plan is executed in the environment.
The wrapper also has a novelty recovery component that can be used to implement routines for novelty recovery. 


\subsubsection{Reward-shaping wrapper}
In complex tasks with a large state space and action spaces, reward shaping is a commonly used technique. With the help of the integrated planner, a filtered list of actions in the PDDL plan gets selected, from which the wrapper generates sub-goals. The user may define the filter criteria and the plan-subgoal correspondences. Through comparison of the state before and after each transition, the wrapper checks whether the subgoal is met and rewards the agent for reaching the sub-goals. The reward-shaping wrapper can help implement sophisticated routines for novelty handling by combining planning and learning.

\subsubsection{External RL agent}
Our architecture also implements a modular reinforcement learning framework (Tianshou~\cite{tianshou}). The availability of this module helps in implementing RL algorithms for training agents. The connectivity of the module with our agent architectures enables sophisticated agent designs and helps users in experimenting with various learning algorithms.

\section{Evaluations}

Evaluating open-world agents necessitates novel protocols. We must also adjust evaluation metrics, considering agent performance both pre- and post-novelty injection.
We define two scenarios,\textit{ pre-novelty} and similarly, \textit{post-novelty} scenario. In the \textit{pre-novelty scenario}, the environment conditions are known to the agent, and the agent's knowledge and/or a pre-trained policy is enough to solve the task \textit{successfully}. However, when a novelty is injected, the agent's knowledge may become incomplete to solve the task either \textit{successfully} (and/or) \textit{optimally}. We call this scenario the \textit{post-novelty scenario}. 
To illustrate, let us consider the introduction of a novel entity axe in the environment (shown in the red box in Figure~\ref{fig:novelgym_representation}). 
In the subsequent sub-sections, we will detail the proposed evaluation protocol, followed by the proposed evaluation metrics.
\subsection{Evaluation Protocol}
The evaluation protocol is divided into four phases:
\paragraph{\textbf{Initial training phase}} In this phase, we train the agent in a controlled environment that is free from novelties. In other words, the agent's knowledge base is complete to solve the task. For a reinforcement learning agent, knowledge can be a shaped reward function, a predefined hierarchical task decomposition~\cite{kulkarni2016hierarchical}, or a Linear Temporal Logic (LTL) guided automaton as a reward function~\cite{icarte2023learning}, etc. For a planning agent, knowledge can be a description of the task through PDDL~\cite{aeronautiques1998pddl}(illustrative example in Table~\ref{tab:symbolic_state}). 
\paragraph{\textbf{Novelty injection}} In this phase, we introduce the novelty into the environment. The introduction of the novelty may hamper the performance of the agent. We can monitor the impact of the performance of the agent immediately upon the introduction of the novelty. Monitoring the impact can help in testing the robustness of the agent in the face of novelties. 
\paragraph{\textbf{Adaptation Phase}} In this phase, we allow the agent to interact with the novel environment to solve the task efficiently. This phase is crucial in gauging how quickly and effectively the agent re-calibrates its approach in response to the introduced novelty.
\paragraph{\textbf{Post-Adaptation evaluation}} After a period of adaptation (predefined time or convergence criteria), we can assess the performance of the agent in the novel environment. 

\subsection{Evaluation Metrics}

\begin{figure}[t]
    \centering
    \includegraphics[width=0.5\textwidth]{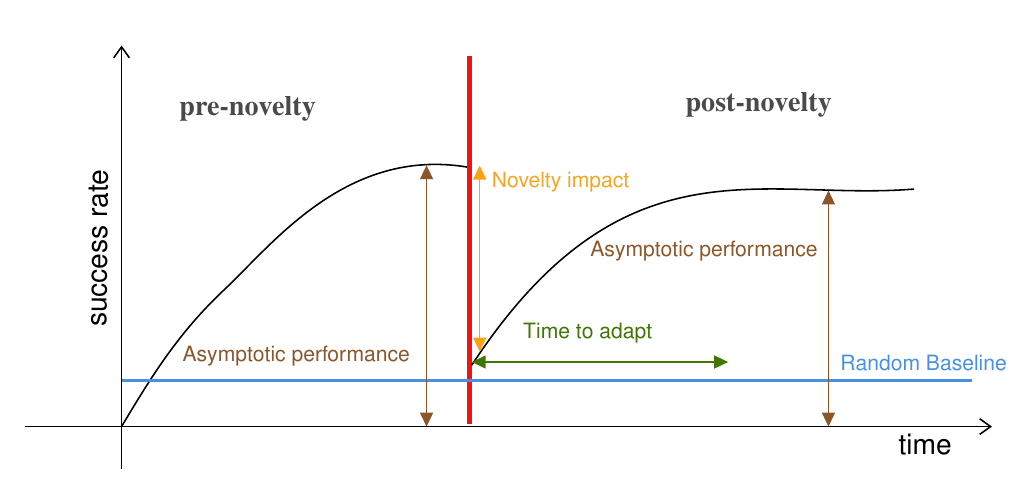}
    \caption{Illustration of performance metrics for open-world agents.}
    \label{fig:eval_metrics}
\end{figure}

We propose five evaluation metrics to evaluate agents in open-world environments, illustrated in Figure~\ref{fig:eval_metrics}. The graph is plotted with respect to the success rates versus time. The \textit{success rate} measures the agent's performance in successfully solving the task. The success rate can be measured at every epoch\footnote{An epoch is the number of episodes/timesteps of training after which we evaluate the agent. In episodic tasks, the agent is provided a quota of timesteps to finish an episode.} 
\textit{Success rate} (\( S \)): Given \( n \) episodes, where \( s \) of them are successful, the success rate is defined as $S = \frac{s}{n}$.
We formally define the evaluation metrics:
\begin{enumerate}
    \item \textbf{Pre-novelty asymptotic performance ($S_{\text{pre-novelty}}$)}: This metric measures the final performance of the agent after the convergence criterion is met in the pre-novelty task (shown in brown in the pre-novelty part of Figure~\ref{fig:eval_metrics}).
    \item \textbf{Novelty impact ($I_{\text{novelty}}$)}: This metric quantifies the immediate effect of introducing novelty on the agent's performance. It is calculated as the difference between the agent's performance before the novelty is introduced and its immediate performance after the novelty is encountered (illustrated in yellow in Figure~\ref{fig:eval_metrics}):
    $$I_{\text{novelty}} = S_{\text{pre-novelty}} - S_{\text{immediate post-novelty}}$$
    If the agent cannot solve the task without further adaptation, its performance can theoretically drop to zero. Alternatively, in certain scenarios, even without immediate adaptation, the agent might still display non-zero performance. 
    \item \textbf{Time to adapt ($T_\text{adapt}$)}: The time taken by the agent to reach the convergence criteria post novelty adaptation is the time to adapt (illustrated in green in Figure~\ref{fig:eval_metrics}). The time taken can be measured in terms of time steps, number of actions taken, or CPU time. 
    \item \textbf{Asymptotic adaptation performance ($S_{\text{post-novelty}}$)}: This metric measures the post-novelty adaptation performance by the agent. This is the success rate in the post-novelty scenario when the convergence criterion is met.
    \item \textbf{Post-Adaptation Efficiency} ($\Delta {t}$): This metric quantifies the agent's policy efficiency after adjusting to novelty relative to its performance before the novelty. Specifically, it captures the potential for beneficial novelties that enable the agent to find task shortcuts. The metric is defined as:
    \vspace{-0.25em}
    $$\Delta t=t_{\text {pre-novelty }}-t_{\text {post-novelty }}$$
    where $t_{\text {pre-novelty }}$ is the average time the agent takes to solve the task before encountering the novelty, and $t_{\text {post-novelty}}$ is the average time post-novelty adaptation. Time measurements can be in time steps, number of actions, or CPU time.
\end{enumerate}


%
\section{Experiments}
\subsection{Task}
\paragraph{\textbf{Pogostick task}} The task is a Minecraft-inspired crafting task as illustrated in the running example (Section~\ref{subsec:running_example}). The goal of the agent is to craft a pogostick while collecting resources and crafting various intermediary items.
We simplified the task to make the task more tractable for a reinforcement learning agent. In the simplified task, the end goal of the agent remains to craft a pogostick while standing in front of the crafting table. The agent starts with a few items already in its inventory, such as a tree tap and iron pickaxe, and the steps required in order to achieve its goal are: (1) approach a block of platinum and (2) break it with the iron pickaxe; (3) approach a block of diamond and (4) break it with the iron pickaxe, (5) craft a plank, then (6) craft a stick, (7) select the tree tap and (8) collect rubber from an oak log while in front of the oak log, (9) approach the trader and (10) trade platinum for titanium, (11) approach the crafting table and (12) craft a block of diamond, and finally (13) craft a pogostick in front of the crafting table.

%
\begin{figure}[t]
    \centering
    \includegraphics[width=0.5\textwidth]{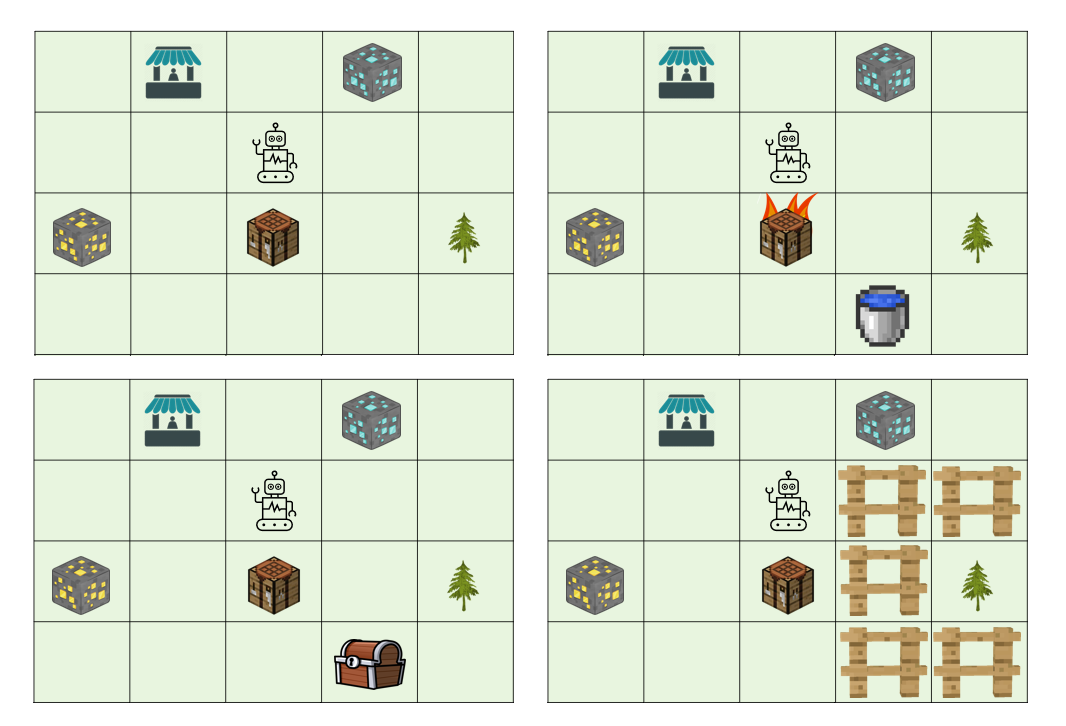}
    \caption{Illustration of the (clockwise) pre-novelty environment, fire novelty, fence novelty and chest novelty.}
    \label{fig:novelties}
\end{figure}
\subsection{Novelties}
We implement a total of 12 transformations of the environment to test algorithms and frameworks for novelty handling in open-world learning\footnote{Tutorial on how an environmental transformation can be implemented is demonstrated in the appendix section I.}. These transformations comprise various aspects of task-solving and robust testing abilities.
\subsubsection{Detrimental}
A novelty in the environment is considered detrimental to an agent if it induces a change that hinders the agent from fulfilling the task for which it was designed. For example, if a tree in the environment can only be broken using a novel entity axe, and the agent has never utilized an axe to break trees. In that case, this transformation is considered detrimental to the agent's task in the environment.
\subsubsection{Beneficial novelties}
A novelty in the environment is considered beneficial for the agent if it induces a change that enhances the agent's ability to solve its task efficiently under a suitable performance metric (e.g. total reward, number of timesteps, etc.).
\subsubsection{Nuisance novelties}
A novelty in the environment is considered nuisance for the agent if it does not affect the agent's task or the agent's representation of the world.

We selected 5 environmental transformations for evaluations~\footnote{A detailed list of environmental transformations and their functionalities is described in the Appendix Section~\ref{sec:novelty_implementations}.}.
\begin{enumerate}
    \item Axe: In this transformation, the tree is unbreakable unless an axe is used to break it. (\textit{Detrimental})
    \item Chest: A chest is placed in the gridworld, and a new action \texttt{approach plastic chest} appears in the agent's action set. If the agent uses the \texttt{collect} action while standing in front of the \textit{chest}, its inventory is filled with all the ingredients necessary to craft a \textit{pogostick}. (\textit{Beneficial})
    \item Trader: One grid cell must be between the agent and a \textit{trader} in order for the agent to be able to execute the \texttt{trade} action with the trader. (\textit{Detrimental})
    \item Fence: All oak logs in the gridworld are surrounded by a \textit{fence} on the neighboring cells. The agent must break the \textit{fence} first in order to access the \textit{oak log}. (\textit{Detrimental})
    \item Fire: The crafting table is set on fire, and a water bucket is placed in the environment. The agent must first collect the water bucket and use it to put out the fire before using the crafting table for any crafting. (\textit{Detrimental})
\end{enumerate}

\subsection{Agent architectures}
In order to develop agent architectures, we adapted a few existing architectures for novelty handling. The architectures ranged from learning approaches to neurosymbolic approaches and were adapted with sophisticated exploration methods. The modular nature of our proposed ecosystem helps in adapting and developing these hybrid architectures. Mainly, we had two approaches transfer learning and hybrid neurosymbolic approach.

\paragraph{Hybrid planning \& learning approach}
The hybrid planning and learning method was a direct implementation of \cite{goel2022rapid}. 
The method assumes the pre-novelty task domain to be defined using PDDL. After novelty injection, if the agent cannot solve the task due to action execution failure, the method instantiates a learning (RL) problem. The goal of the RL problem is to find a plannable state. The plannable state is either a state that satisfies the failed operator's effects or can help the agent jump ahead in the plan. The instantiated learning problem can be solved by any off-the-shelf RL algorithm. We used PPO~\cite{schulman2017proximal} as the RL agent. 
We also adapted the Intrinsic Curiosity Module (ICM)~\cite{pathak2017curiosity} for robust exploration.
\begin{table*}[t]
\small
    \centering
    \begin{tabular}{ll ccccc}
        \toprule
         Novelty & Agent & $S_{\text{pre-novelty}} \uparrow $ & $I_{\text{novelty}} \downarrow $  & $T_\text{adapt} \downarrow$ & $S_{\text{post-novelty}}\uparrow$ & $\Delta t \downarrow$\\
         \midrule
         \textbf{Axe} & RapidLearn$^{+}$(PPO+ICM) & $1 \pm 0$ & $ 0.83 \pm 0.205 $ & $ 71040 \pm 17413 $ & $ 1.0 \pm 0.0 $ & $ 122.7 \pm 15.61 $ \\
                        & RapidLearn (PPO)  & $1 \pm 0$ & $ 0.69 \pm 0.207 $ & $ 68160 \pm 17934 $ & $ 1.0 \pm 0.0 $ & $ 116.9 \pm 9.88 $ \\
                        & Transfer RL (PPO)+ICM & $ 1 \pm 0 $ & $ 1.0 \pm 0.0 $ & $ 169440 \pm 97949 $ & $ 0.97 \pm 0.021 $ & $ 61.0 \pm 8.95 $ \\
                        & Transfer RL (PPO) & $1 \pm 0$ & $ 1.0 \pm 0.0 $ & $ 114240 \pm 19651 $ & $ 0.97 \pm 0.018 $ & $ 63.1 \pm 8.17 $ \\
        \midrule
        \textbf{Chest} & RapidLearn$^{+}$(PPO+ICM) & $1 \pm 0$ & -- & -- & -- & -- \\
                        & RapidLearn (PPO)  & $1 \pm 0$ & -- & -- & -- & -- \\
                        & Transfer RL (PPO)+ICM & $ 1 \pm 0 $ & $ 0.0 \pm 0.004 $ & $ 24000 \pm 0 $ & $ 1.0 \pm 0.0 $ & $ -3.0 \pm 0.78 $ \\
                        & Transfer RL (PPO) & $1 \pm 0$ & $ 0.01 \pm 0.009 $ & $ 24000 \pm 0 $ & $ 0.98 \pm 0.011 $ & $ -3.5 \pm 3.13 $ \\
        \midrule
        \textbf{Trader} & RapidLearn$^{+}$(PPO+ICM) & $1 \pm 0$ & $ 0.45 \pm 0.069 $ & $ 102720 \pm 20724 $ & $ 0.95 \pm 0.017 $ & $ 126.9 \pm 9.97 $ \\
                        & RapidLearn (PPO)  & $1 \pm 0$ & $ 0.5 \pm 0.07 $ & $ 96480 \pm 21191 $ & $ 0.96 \pm 0.02 $ & $ 122.8 \pm 8.93 $ \\
                        & Transfer RL (PPO)+ICM & $ 1 \pm 0 $ & $ 0.96 \pm 0.108 $ & $ 227040 \pm 112120 $ & $ 0.99 \pm 0.011 $ & $ 93.6 \pm 12.05 $ \\
                        & Transfer RL (PPO) & $1 \pm 0$ & $ 0.94 \pm 0.128 $ & $ 350880 \pm 237039 $ & $ 0.96 \pm 0.025 $ & $ 89.3 \pm 7.87 $ \\
        \midrule
        \textbf{Fence} &  RapidLearn$^{+}$(PPO+ICM) & $1 \pm 0$ & $ 0.38 \pm 0.102 $ & $ 69120 \pm 15657 $ & $ 0.96 \pm 0.02 $ & $ 161.6 \pm 7.6 $ \\
                        & RapidLearn (PPO)  & $1 \pm 0$ & $ 0.41 \pm 0.052 $ & $ 66400 \pm 20999 $ & $ 0.95 \pm 0.008 $ & $ 168.7 \pm 6.15 $ \\
                        & Transfer RL (PPO)+ICM & $ 1 \pm 0 $ & $ 1.0 \pm 0.0 $ & $ 854400 \pm 52974 $ & $ 0.93 \pm 0.02 $ & $ 73.4 \pm 4.7 $ \\
                        & Transfer RL (PPO) & $1 \pm 0$ & $ 1.0 \pm 0.0 $ & $ 701760 \pm 100589 $ & $ 0.92 \pm 0.024 $ & $ 90.9 \pm 9.49 $ \\
        \midrule
        \textbf{Fire} &  RapidLearn$^{+}$(PPO+ICM) & $1 \pm 0$ & $ 1.0 \pm 0.0 $ & $ 263520 \pm 81979 $ & $ 0.95 \pm 0.02 $ & $ 133.2 \pm 21.38 $ \\
                        & RapidLearn (PPO)  & $1 \pm 0$ & $ 1.0 \pm 0.0 $ & $ 252000 \pm 49663 $ & $ 0.95 \pm 0.016 $ & $ 142.5 \pm 13.71 $ \\
                        & Transfer RL (PPO)+ICM & $ 1 \pm 0 $ & $ 1.0 \pm 0.0 $ & $ 372960 \pm 198096 $ & $ 0.94 \pm 0.016 $ & $ 105.3 \pm 9.85 $ \\
                        & Transfer RL (PPO) & $1 \pm 0$ & $ 1.0 \pm 0.0 $ & $ 127200 \pm 37932 $ & $ 0.96 \pm 0.029 $ & $ 103.3 \pm 13.22 $ \\
         \bottomrule
    \end{tabular}
    \caption{Evaluations of the agents across five novelty scenarios. $\uparrow$ indicates higher value is better, $\downarrow$ indicated lower value is better and 
    -- indicates that the evaluated novelty is not a novelty for the corresponding agent.}
    \label{tab:results}
\end{table*}
The modularity of NovelGym played a crucial role in the easy implementation of such a complicated agent architecture.

\paragraph{Transfer RL}
Due to the complex nature of the task, it was challenging to train an RL agent to solve it. We used a dense reward function to train the RL agent for the pre-novelty task~\footnote{Details about the reward function are in the appendix Section~\ref{sec:reward_function}.}. The RL agent was trained for $4$ Million timesteps after which it achieved comparable performance to a planning agent. In the case of novelty injection, the agent's action space and observation space were expanded, and the weights of the pre-novelty task were transferred. Our architecture's flexibility helped us achieve automatic expansion on-the-fly.
The agent architectures we benchmark are:
\begin{itemize}
    \item RapidLearn$^{+}$(PPO+ICM): Hybrid planning and learning using PPO as the RL algorithm integrated with ICM.
    \item RapidLearn(PPO): Hybrid planning and learning with PPO.
    \item Transfer RL(PPO)+ICM: Transfer learning with ICM.
    \item Transfer RL (PPO): Transfer learning using PPO.
\end{itemize}
The agent's subsymbolic observation space is a LIDAR-like sensor that emits beams for every entity in the environment at incremental angles of $\frac{\pi}{4}$ to determine the closest entity in the angle of beam (similar illustration shown in Figure~\ref{fig:sensor_representation} (left)).
The size of lidar sensor observation is $8 \times |\mathcal{E}|$, where $\mathcal{E}$ is the set of entities in the environment. The observation space is augmented by additional sensors that observe the agent's contents of inventory and entity selected by the agent. 
The agent's symbolic description was represented using PDDL, where entities were described with a type hierarchy, and predicates represented the relations between the types. The actions were described using preconditions and effects~\footnote{A detailed PDDL is added in Appendix Section~\ref{sec:pddl}.}.

We also implement an image-based local view representation. The local view representation is a one-hot vector encoding of all the entity types in the world in a grid of $n \times n$ around the agent~\footnote{More details in Appendix Section~\ref{sec:observation_spaces}.}.

The agent has primitive actions such as move-forward, turn-left, etc., and parameterized actions such as approach<entity>, select <entity>, etc.~\footnote{A detailed list of actions can be found in the Appendix Section~\ref{sec:action_space}.}. The action space size for the pre-novelty task is 28.

\section{Results \& Discussion}
The results of our experiments, which tested five different novelties on four agents, are detailed in Table~\ref{tab:results}~\footnote{Runs were averaged across 10 random seeds.}. Listed in the table are the mean and standard deviation of the metrics across all seeds. Notably, the metric \( S_{\text{pre-novelty}} \) consistently scored perfectly. To achieve this, the RL agents underwent training of 4 million timesteps.
Upon introducing novelty, there was a sharp performance decline across all agents for every novelty type, with the exception being the ``chest'' novelty. The chest novelty, being \textit{beneficial} to the agent in the given environment, does not interfere with the task. Hence, the agent's success rate remains unaffected. The hybrid agent, given its design, does not adapt to novelties unless an execution failure is detected. Therefore, for the hybrid agent, this is not a novelty. The \( \Delta t \) metric (lower the better), which measures adaptation efficiency, highlights the superior performance of the transfer RL agents in adapting to ``chest" novelty. This performance can be attributed to the exploratory nature of the RL algorithms.
The \( I_{\text{novelty}} \) metric (lower the better) was computed by measuring the performance of each agent before training for adaptation for the novelty. Its values reveal that learning-based strategies are more sensitive to novelty introduction. 
However, hybrid models can sometimes exhibit marginally superior immediate adaptation upon the introduction of novelty. This might be due to their targeted novelty adaptation approach. We can observe from the results that fire novelty resulted in the sharpest decline in the performance of all the agents. This shows that the adaptation of this novelty based on the agents we evaluated is relatively harder than others. The correspondence of the \( I_{\text{novelty}} \) metric to the higher values of \( T_{\text{adapt}} \) can be observed in all the novelty and agents cases. 
For all the novelties, hybrid agents surpassed transfer learning methods in the \( T_{\text{adapt}} \) metric. This superiority may arise from hybrid architectures' focused learning and their ability to effectively reuse knowledge. Furthermore, we observed that the inclusion of ICM enhances adaptation in some cases. However, in some cases, we can see that the ICM approach, especially when adapted to the transfer learning agent, deteriorates the performance (fire novelty). The variance in case of chest novelty in the \( T_{\text{adapt}} \) is 0 because the agent did not achieve a low success rate to adapt and therefore it was not trained based on a convergence but rather a fixed set of episodes.
$S_\text{post-novelty}$ metric results do not show perfect scores for some agents. This is because the agents were trained to satisfy a convergence criteria. They may reach perfect scores if trained longer. 

Our results highlight essential elements for crafting AI agents capable of adeptly handling novelties. Specifically, the $T_\text{adapt}$ metric reveals that even with advanced hybrid methods, a significant number of environmental interactions are needed for agents to re-calibrate to their original performance levels after facing novelty. In real-world scenarios, especially in robotics, the opportunity for such extensive interactions is limited. This underscores the importance of our current domain and architectural approach, positioning it as a promising avenue for furthering open-world learning research. 
\section{Conclusion \& Future Work}

We introduced NovelGym, a flexible platform tailored for the implementation and injection of novelties and easy task creation in gridworld environments. We highlighted the modularity of our proposed ecosystem, substantiated by the integration of multiple novelties and the implementation of complex agent architectures. To further support the evaluation of novelty-aware open-world agents, we proposed an evaluation protocol complemented by evaluation metrics. Our empirical results offer insights into the performance dynamics of various agent architectures in the face of multiple novelties, evaluated against the backdrop of our introduced metrics. Our benchmarking data also presents the intricacies and challenges to open-world learning. 
Building on the strengths of NovelGym, the platform's adaptable and robust architecture presents a ground for many research directions. A possible direction can be the procedural and automatic generation of novelties, paired with an emphasis on continual agent learning. 
As a part of future work, we aim to expand the capabilities of NovelGym to fully embrace multi-agent dynamics for deeper collaborative and competitive learning paradigms. We would also like to extend the environment to support human-in-the-loop learning, as novelty handling could be potentially benefited by demonstrations through humans and incorporating those into learning.



.

\bibliographystyle{ACM-Reference-Format} 
\bibliography{root}


\clearpage
\section*{Appendix}
\appendix

\section{Convergence Criteria} 
The learner is considered to have converged if the following conditions are met:
    \begin{itemize}
        \item The agent achieves the goal \( \delta g \geq \delta G \) in the last \( \eta \) epochs.
        \item The average reward in these epochs is \( \delta r \geq \delta R \).
        \item The max success rate and the max reward in the past \( \eta + \upsilon \) epochs is not greater to the those in the last \( \eta \) epochs.
    \end{itemize}
    where each epoch is 4800 time steps, $\eta = 5$, $\upsilon = 5$, $\delta G = 0.9$, $\delta R = 400$. The episode length is 400 time steps. Episode ends if the agent reaches the goal state or if it completes 400 time steps.
\section{Action Space}
\label{sec:action_space}
The action space for the hybrid agent is composed of higher level actions operators such as \texttt{approach <entity>},etc. These are implemented as the executors that use the primitive actions available in the environment to execute the operators. Therefore the hybrid agent's action space has all the actions including higher level operarators (implemented as lower level action executors) as well as primitive actions. For the RL agent, we have each higher level operator that is grounded. The exhaustive list of the actions we used:
\begin{itemize}
    \item collect
    \item break\_block
    \item approach\_oak\_log
    \item approach\_diamond\_ore
    \item approach\_crafting\_table
    \item approach\_block\_of\_platinum
    \item approach\_entity\_103
    \item interact\_103
    \item select\_oak\_log
    \item select\_iron\_pickaxe
    \item select\_sapling
    \item select\_tree\_tap
    \item select\_crafting\_table
    \item deselect\_item
    \item craft\_stick
    \item craft\_planks
    \item craft\_block\_of\_diamond
    \item craft\_pogo\_stick
    \item trade\_block\_of\_titanium\_1
    \item move\_forward
    \item move\_backward
    \item move\_left
    \item move\_right
    \item rotate\_left
    \item rotate\_right
    \item place
    \item <...>
    \item <...>
\end{itemize}
where the last two slots varies during post-novelty situations.

\section{Observation Spaces}
\label{sec:observation_spaces}
\subsection{LiDAR Representation}

In this representation, the agent's observation space is a gymnasium (openAI Gym) Box space. This Box space is made from a one-dimensional vector created by concatenating three vectors for the world as observed by the agent, the agent's inventory, and the item selected by the agent. The agent's world observation is generated in the following way: 8 beams of 45 degrees each are sent from 0 to 360 degrees, and the euclidean distances of the objects that strike the LiDAR are stored in a two-dimensional array whose rows correspond to the integer encodings of all the possible objects and entities in the environment and whose columns correspond to the individual directions in which a LiDAR beam is sent. The assumption is made that occlusions do not hold true. The two-dimensional array is subsequently flattened. The inventory is represented as a one-dimensional array where the integer at each index corresponds to the number of objects encoded by the index that the agent has. The selected item is represented as a one-hot encoding in a one-dimensional array. Objects and entities in all three cases are assigned integer encodings. The low and high bounds of the Box space are arrays of a length corresponding to the maximum number of object and entity types in the environment and of maximum values corresponding to the maximum count for each object and entity type. The size of the observation used during our experiment is 250, which includes placeholders for two additional objects in post-novelty situations. We added placeholders for convenience in implementation. In principle our architecture can automatically expand the network based on new entities.

\subsection{Image-based Representation}

In this representation, the observation space of the agent is a dictionary with three key-value pairs: for the agent's local view, for the agent's inventory, and for the agent's selected item. Each of the values in the dictionary is a gymnasium Box space, where the low and high bounds are given by the maximum number of instances of an object or entity in the environment. The agent's local view is represented as a two-dimensional square image, where the number of one-hot encoded channels corresponds to the number of possible object and entity types in the environment. The inventory is represented as a one-dimensional array where the integer at each index corresponds to the number of objects encoded by the index that the agent has. The selected item is represented as a one-hot encoding in a one-dimensional array. Objects and entities in all three cases are represented by integer encodings.

\section{Reward Function}
\label{sec:reward_function}

\begin{table*}[ht!]
    \centering
    \small
    \begin{tabular} {p{0.12\linewidth} p{0.12\linewidth} p{0.68\linewidth}}
    \toprule
    
    \textbf{Novelty} & \textbf{Category}  & \textbf{Description} \tabularnewline  \arrayrulecolor{taupegray} \midrule
    
    \emph{Axe} & \emph{Detrimental} & The agent is provided with an axe in its inventory at the start of the episode, and a new action ``select axe'' is added to its action set. The agent cannot break a tree unless it has selected the axe beforehand. \tabularnewline
    \arrayrulecolor{taupegray} \midrule
    
    \emph{Busy} & \emph{Nuisance} & Each trader is busy during a certain percentage of time steps. During time steps that a trader is busy, the agent cannot successfully execute a trade against them. \tabularnewline
    \arrayrulecolor{taupegray} \midrule
    
    \emph{Chest} & \emph{Beneficial} & A plastic chest containing all the ingredients required for crafting a pogostick appears in the grid world, and a new action ``approach plastic chest'' is added to the agent's action set. The chest can be acted upon with the ``collect'' action, upon which the ingredients contained in the chest are added to the agent's inventory.  \tabularnewline
    \arrayrulecolor{taupegray} \midrule
    
    \emph{Distance} & \emph{Detrimental} & One cell must separate the trader and the agent in order for the trade action to execute successfully. \tabularnewline
    \arrayrulecolor{taupegray} \midrule
    
    \emph{Fence} & \emph{Detrimental} & A breakable fence surrounds each tree in the grid world, and a new action ``approach fence'' is added to the agent's action set. \tabularnewline
    \arrayrulecolor{taupegray} \midrule
    
    \emph{Fire} & \emph{Detrimental} & The crafting table in the grid world is given a new property ``on fire'', and this property is set to true at the beginning of the episode. Somewhere else in the grid world, a water bucket is placed. The agent can extinguish the fire by choosing either the ``use'' action or the ``collect'' action while near the crafting table and holding the water bucket, whereafter it is returned a ``bucket'' in its inventory. The crafting table can only be crafted at after the fire has been extinguished. \tabularnewline
    \arrayrulecolor{taupegray} \midrule
    
    \emph{Moving} & \emph{Nuisance} & The traders are given action sets with ``move forward'', ``move backward'', ``move left'', and ``move right'' and select one of these actions randomly at every time step. \tabularnewline
    \arrayrulecolor{taupegray} \midrule
    
    \emph{Multi-interact} & \emph{Detrimental} & There are two variants of this novelty. Either trees cannot be broken with the ``break'' action or there are no trees in the grid world at all. In either case, the agent can obtain oak logs by interacting with the traders. \tabularnewline
    \arrayrulecolor{taupegray} \midrule
    
    \emph{Multi-room} & \emph{Nuisance} & Two rooms separated by a wall and doorway comprise the grid world rather than just one large room. \tabularnewline
    \arrayrulecolor{taupegray} \midrule
    
    \emph{Portal Treasure} & \emph{Detrimental} & The agent is given a treasure in its inventory, a portal is placed in the grid world, and no tree can be broken using the break action. If the agent selects the action ``use'' while at the portal and with the treasure in its inventory, it loses the treasure but receives a number of oak logs in its inventory instead. \tabularnewline
    \arrayrulecolor{taupegray} \midrule
    
    \emph{Random drop} & \emph{Detrimental} & The agent can no longer break trees using the break action nor any other action. However, with a certain probability, the action ``break block'' results in a given number of oak logs being dropped into the grid world. \tabularnewline
    \arrayrulecolor{taupegray} \midrule
    
    \emph{Space Around} & \emph{Detrimental} & At the beginning of the episode, the agent is given a number of saplings in its inventory. These saplings can be placed in the grid world and grow into trees. The agent cannot place a sapling when there is a tree or wall within radius 1 of the item. \tabularnewline
    \arrayrulecolor{taupegray} \bottomrule
    
    \end{tabular}
    \caption{Novelties currently implemented in the NovelGym environment}
    \label{tab:rl_systemic_examples}
\end{table*}
\begin{algorithm}
\caption{Training Routine with Crafted Reward Function}
\label{alg:reward_func}
\begin{algorithmic}[1]
\State Plan $P$
\State Ordered set (queue) $\mathcal{Q}$ of subgoals derived from $P$
\State Set of past goals $\mathcal{G}_{\text{past}} \gets \emptyset$

\While{$\mathcal{Q}$ is not empty}
    \State Let $g_{\text{current}}$ be the first element of $\mathcal{Q}$
    
    \If{$g_{\text{current}}$ is achieved}
        \State $\mathcal{G}_{\text{past}} \gets \mathcal{G}_{\text{past}} \cup \{ g_{\text{current}} \}$
        \State Remove $g_{\text{current}}$ from $\mathcal{Q}$
        
        \If{$g_{\text{current}}$ was the immediate predecessor in $\mathcal{G}_{\text{past}}$}
            \State $\text{Reward}(g_{\text{current}}) \gets \frac{1}{2} \times \text{Reward}(g_{\text{current}})$
        \EndIf
        
        \For{each $g_{\text{subsequent}}$ in $\mathcal{Q}$}
            \If{$g_{\text{subsequent}}$ is achieved}
                \State Remove $g_{\text{subsequent}}$ from $\mathcal{Q}$
                \State $\text{Reward}(g_{\text{subsequent}}) \gets 0$
            \Else
                \State \textbf{break}
            \EndIf
        \EndFor
    \Else
        \State Update model with $\text{Reward}(g_{\text{current}})$
    \EndIf
\EndWhile
\end{algorithmic}
\end{algorithm}
\subsection{Pre-Novelty Base RL}
Algorithm for training with a crafted reward function is shown in Algorithm~\ref{alg:reward_func}.
Initially, a plan is created using the planner, and then the plan is filtered and given rewards for the following subgoals:

\begin{itemize}
    \item \texttt{break <block of platinum>},
    \item \texttt{trade block of titanium},
    \item \texttt{break block of diamond},
    \item \texttt{break tree},
    \item \texttt{craft planks},
    \item \texttt{craft stick},
    \item \texttt{craft pogostick}.
\end{itemize}

Note that the order of these subgoals is also considered in the assessment, and a reward will only be given for the first subgoal in the queue. Once a subgoal is achieved for the first time, it becomes a ``past goal''. Whenever the immediately preceding ``past goal'' is accomplished, the reward given for this subgoal is decayed by half. Additionally, whenever a certain subgoal is achieved, the environment will check for the subsequent subgoals in the queue and compare these against the agent's inventory. If this inventory already matches the outcome of any of these subgoals, the given subgoal is considered completed, with no reward issued, and the subgoal after it becomes the next subgoal in the queue.

\subsection{Post-Novelty Transfer RL}

A reward of 1000 is given for achieving the goal of crafting a pogostick, else a negative reward of -1 is issued for each time step.

\subsection{Post-Novelty Hybrid Planning and Learning}

A reward of 1000 is given for crafting a pogostick, a negative reward of -1 is issued with each timestep, and a negative reward of -250 is a result of the agent getting into an unplannable state.

\section{Novelty Implementations}
\label{sec:novelty_implementations}

For a list of the environment trandformations currently implemented in the NovelGym environment, see Table~\ref{tab:rl_systemic_examples}. The components of a novelty being integrated into the system are as follows.

\begin{itemize}
    \item \textbf{YAML config file}: This config file serves as an add-on to the existing YAML config file that configures the environment. Anything included in the new config expands on the settings in the environment config, and anything included under the key ``novelties'' overrides any chosen component of the original settings. Common entries to the config add or modify actions, action sets, action preconditions and effects, entities, entity properties, and the grid world layout. A novelty can be created through a YAML config file alone when it does not involve adding or modifying any modules, an example of which are the \textit{Multi-room} and \textit{Distance} novelties in Table~\ref{tab:rl_systemic_examples}. This is possible when no new source Python module is required for the novelty to function.
    \item \textbf{Python source file(s)}: In the case that a novelty requires the addition or modification of an entity or action module, it is necessary to include the Python source code for this adjustment. All the novelties specified in Table~\ref{tab:rl_systemic_examples} except for \textit{Multi-room} and \textit{Distance} involve the addition of one or more modules. Note that changing the details of an existing module can be accomplished by changing the source Python code for the module in the novelty config file.
\end{itemize}

\section{Sample Configuration File}
Here we provide a sample configuration file, which generates trees in the world, allows the user to move around in the world, break the trees, and craft planks from the trees:
\begin{lstlisting}
---
actions:
  break_block:
    module: gym_novel_gridworlds2.contrib.polycraft.actions.Break
    step_cost: 3600
  move_forward:
    module: gym_novel_gridworlds2.contrib.polycraft.actions.SmoothMove
    direction: W
  move_backward:
    module: gym_novel_gridworlds2.contrib.polycraft.actions.SmoothMove
    direction: X
  move_left:
    module: gym_novel_gridworlds2.contrib.polycraft.actions.SmoothMove
    direction: A
  move_right:
    module: gym_novel_gridworlds2.contrib.polycraft.actions.SmoothMove
    direction: D
  craft:
    module: gym_novel_gridworlds2.contrib.polycraft.actions.Craft
  trade:
    module: gym_novel_gridworlds2.contrib.polycraft.actions.Trade
action_sets:
  main:
  - break_block
  - craft_planks
  - move_*
object_types:
  default: gym_novel_gridworlds2.contrib.polycraft.objects.PolycraftObject
  oak_log: gym_novel_gridworlds2.contrib.polycraft.objects.easy_oak_log.OakLog
map_size: [16, 16]
rooms:
  '2':
    start: [0, 0]
    end: [15, 15]
objects:
  oak_log:
    quantity: 5
    room: 2
    chunked: 'False'
entities:
  main_1:
    agent: gym_novel_gridworlds2.agents.KeyboardAgent
    name: entity.polycraft.Player.name
    type: agent
    entity: gym_novel_gridworlds2.contrib.polycraft.objects.PolycraftEntity
    action_set: main
    inventory:
      iron_pickaxe: 1
      tree_tap: 1
    id: 0
    room: 2
    max_step_cost: 100000
recipes:
  planks:
    input:
    - oak_log
    - '0'
    - '0'
    - '0'
    output:
      planks: 4
    step_cost: 1200
trades: {}
auto_pickup_agents:
- 0

\end{lstlisting}

\section{PDDL}
\label{sec:pddl}
\subsection{Pre-Novelty Domain}
The pre novelty domain is defined as followed:
\begin{lstlisting}
(define (domain polycraft_generated)

(:requirements :typing :strips :fluents :negative-preconditions :equality)

(:types 
    pickaxe_breakable - breakable
    hand_breakable - pickaxe_breakable
    breakable - placeable
    placeable - physobj
    physobj - physical
    actor - physobj
    trader - actor
    pogoist - actor
    agent - actor
    oak_log - log
    distance - var
    agent - placeable
    trader - placeable
    pogoist - placeable
    bedrock - placeable
    door - placeable
    safe - placeable
    plastic_chest - placeable
    tree_tap - placeable
    oak_log - hand_breakable
    diamond_ore - pickaxe_breakable
    iron_pickaxe - physobj
    crafting_table - placeable
    block_of_platinum - pickaxe_breakable
    block_of_titanium - placeable
    sapling - placeable
    planks - physobj
    stick - physobj
    diamond - physobj
    block_of_diamond - physobj
    rubber - physobj
    pogo_stick - physobj
    blue_key - physobj
)

(:constants 
    air - physobj
    one - distance
    two - distance
    rubber - physobj
    blue_key - physobj
)

(:predicates ;todo: define predicates here
    (holding ?v0 - physobj)
    (floating ?v0 - physobj)
    (facing_obj ?v0 - physobj ?d - distance)
    (next_to ?v0 - physobj ?v1 - physobj)
)


(:functions ;todo: define numeric functions here
    (world ?v0 - physobj)
    (inventory ?v0 - physobj)
    (container ?v0 - physobj ?v1 - physobj)
)

; define actions here
(:action approach
    :parameters    (?physobj01 - physobj ?physobj02 - physobj )
    :precondition  (and
        (>= ( world ?physobj02) 1)
        (facing_obj ?physobj01 one)
    )
    :effect  (and
        (facing_obj ?physobj02 one)
        (not (facing_obj ?physobj01 one))
    )
)

(:action approach_actor
    :parameters    (?physobj01 - physobj ?physobj02 - actor )
    :precondition  (and
        (facing_obj ?physobj01 one)
    )
    :effect  (and
        (facing_obj ?physobj02 one)
        (not (facing_obj ?physobj01 one))
    )
)

(:action break
    :parameters    (?physobj - hand_breakable)
    :precondition  (and
        (facing_obj ?physobj one)
        (not (floating ?physobj))
    )
    :effect  (and
        (facing_obj air one)
        (not (facing_obj ?physobj one))
        (increase ( inventory ?physobj) 1)
        (increase ( world air) 1)
        (decrease ( world ?physobj) 1)
    )
)


(:action break_holding_iron_pickaxe
    :parameters    (?physobj - pickaxe_breakable ?iron_pickaxe - iron_pickaxe)
    :precondition  (and
        (facing_obj ?physobj one)
        (not (floating ?physobj))
        (holding ?iron_pickaxe)
    )
    :effect  (and
        (facing_obj air one)
        (not (facing_obj ?physobj one))
        (increase ( inventory ?physobj) 1)
        (increase ( world air) 1)
        (decrease ( world ?physobj) 1)
    )
)

(:action break_diamond_ore
    :parameters    (?iron_pickaxe - iron_pickaxe)
    :precondition  (and
        (facing_obj diamond_ore one)
        (not (floating diamond_ore))
        (holding ?iron_pickaxe)
    )
    :effect  (and
        (facing_obj air one)
        (not (facing_obj diamond_ore one))
        (increase ( inventory diamond) 9)
        (increase ( world air) 1)
        (decrease ( world diamond_ore) 1)
    )
)

(:action select
    :parameters    (?prev_holding - physobj ?obj_to_select - physobj)
    :precondition  (and
        (>= ( inventory ?obj_to_select) 1)
        (holding ?prev_holding)
        (not (= ?obj_to_select air))
    )
    :effect  (and
        (holding ?obj_to_select)
        (not (holding ?prev_holding))
    )
)

(:action deselect_item
    :parameters    (?physobj01 - physobj)
    :precondition  (and
        (holding ?physobj01)
        (not (holding air))
    )
    :effect  (and
        (not (holding ?physobj01))
        (holding air)
    )
)

(:action place_sapling
    :parameters (?sapling - sapling ?log - log)
    :precondition (and
        (facing_obj air one)
        (holding ?sapling)
    )
    :effect (and
        (facing_obj ?log one)
        (not (facing_obj air one))
        (increase ( world ?log) 1)
        (decrease ( inventory ?log) 1)
    )
)

(:action place
    :parameters   (?physobj01 - placeable)
    :precondition (and
        (facing_obj air one)
        (holding ?physobj01)
    )
    :effect (and 
        (facing_obj ?physobj01 one)
        (not (facing_obj air one))
        (increase ( world ?physobj01) 1)
        (decrease ( inventory ?physobj01) 1)
    )
)


(:action collect_from_tree_tap
    :parameters (?actor - actor ?log - log)

    :precondition (and
        (holding tree_tap)
        (facing_obj ?log one)
    )
    :effect (and
        (increase ( inventory rubber) 1)
    )
)

; additional actions, including craft and trade
(:action craft_stick
    :parameters ()
    :precondition (and
        (>= ( inventory planks) 2)
    )
    :effect (and
        (decrease ( inventory planks) 2)
        (increase ( inventory stick) 4)
    )
)


(:action craft_planks
    :parameters ()
    :precondition (and
        (>= ( inventory oak_log) 1)
    )
    :effect (and
        (decrease ( inventory oak_log) 1)
        (increase ( inventory planks) 4)
    )
)


(:action craft_block_of_diamond
    :parameters ()
    :precondition (and
        (facing_obj crafting_table one)
        (>= ( inventory diamond) 9)
    )
    :effect (and
        (decrease ( inventory diamond) 9)
        (increase ( inventory block_of_diamond) 1)
    )
)


(:action craft_tree_tap
    :parameters ()
    :precondition (and
        (facing_obj crafting_table one)
        (>= ( inventory planks) 5)
        (>= ( inventory stick) 1)
    )
    :effect (and
        (decrease ( inventory planks) 5)
        (decrease ( inventory stick) 1)
        (increase ( inventory tree_tap) 1)
    )
)


(:action craft_pogo_stick
    :parameters ()
    :precondition (and
        (facing_obj crafting_table one)
        (>= ( inventory stick) 2)
        (>= ( inventory block_of_titanium) 1)
        (>= ( inventory diamond) 1)
        (>= ( inventory rubber) 1)
    )
    :effect (and
        (decrease ( inventory stick) 2)
        (decrease ( inventory block_of_titanium) 1)
        (decrease ( inventory diamond) 1)
        (decrease ( inventory rubber) 1)
        (increase ( inventory pogo_stick) 1)
    )
)


(:action trade_block_of_titanium_1
    :parameters ()
    :precondition (and
        (facing_obj entity_103 one)
        (>= ( inventory block_of_platinum) 1)
    )
    :effect (and
        (decrease ( inventory block_of_platinum) 1)
        (increase ( inventory block_of_titanium) 1)
    )
)

)


\end{lstlisting}

\subsection{Pre-Novelty Problem}
The pre novelty problem is defined as followed:
\begin{lstlisting}
(define (problem polycraft_problem)
(:domain polycraft_generated)
    (:objects 
        agent - agent
        trader - trader
        pogoist - pogoist
        bedrock - bedrock
        door - door
        safe - safe
        plastic_chest - plastic_chest
        tree_tap - tree_tap
        oak_log - oak_log
        diamond_ore - diamond_ore
        iron_pickaxe - iron_pickaxe
        crafting_table - crafting_table
        block_of_platinum - block_of_platinum
        block_of_titanium - block_of_titanium
        sapling - sapling
        planks - planks
        stick - stick
        diamond - diamond
        block_of_diamond - block_of_diamond
        rubber - rubber
        pogo_stick - pogo_stick
        blue_key - blue_key
        entity_0 - agent
        entity_103 - trader
        entity_102 - pogoist
    )

    (:init
        (= (world air) 178)
        (= (world bedrock) 60)
        (= (world crafting_table) 1)
        (= (world entity_102) 1)
        (= (world oak_log) 5)
        (= (world entity_0) 1)
        (= (world entity_103) 1)
        (= (world diamond_ore) 4)
        (= (world plastic_chest) 1)
        (= (world block_of_platinum) 4)
        (= (world agent) 0)
        (= (world trader) 0)
        (= (world pogoist) 0)
        (= (world door) 0)
        (= (world safe) 0)
        (= (world tree_tap) 0)
        (= (world block_of_titanium) 0)
        (= (world sapling) 0)
        (= (inventory iron_pickaxe) 1)
        (= (inventory tree_tap) 1)
        (= (inventory agent) 0)
        (= (inventory trader) 0)
        (= (inventory pogoist) 0)
        (= (inventory bedrock) 0)
        (= (inventory door) 0)
        (= (inventory safe) 0)
        (= (inventory plastic_chest) 0)
        (= (inventory oak_log) 0)
        (= (inventory diamond_ore) 0)
        (= (inventory crafting_table) 0)
        (= (inventory block_of_platinum) 0)
        (= (inventory block_of_titanium) 0)
        (= (inventory sapling) 0)
        (= (inventory planks) 0)
        (= (inventory stick) 0)
        (= (inventory diamond) 0)
        (= (inventory block_of_diamond) 0)
        (= (inventory rubber) 0)
        (= (inventory pogo_stick) 0)
        (= (inventory blue_key) 0)
        (facing_obj air one)
        (holding air)
    )

    (:goal (>= (inventory pogo_stick) 1))
)

\end{lstlisting}

\vspace{4em}

\section{Hyperparameters}
\label{sec:hyperparameters}
\subsection{PPO}
Table~\ref{tab:ppo-param} shows the hyperparameters for the PPO algorithm:
\begin{table}[htbp]
    \centering
    \begin{tabular}{cc}
    \toprule
        Name &  Value  \\
    \midrule
        Hidden Dense NN Layers  & 256, 64\\
        Optimizer & Adam \\
        Learning Rate & $10^{-5}$ \\
        $\gamma$ (discount factor) & 0.99 \\
        $\epsilon$ (PPO clip) & 0.2 \\
    \bottomrule
    \end{tabular}
    \caption{PPO hyperparameters}
    \label{tab:ppo-param}
\end{table}

\subsection{ICM}
Table~\ref{tab:icm-param} shows the hyperparameters for the Intrinsic Curiosity module:
\begin{table}[htbp]
    \centering
    \begin{tabular}{cc}
    \toprule
        Name &  Value  \\
    \midrule
        Feature Net Hidden Layers & 256, 64 \\
        Feature Dimension & 16 \\
        Learning Rate Scale & 1 \\
        Reward Scale & 0.01 \\
        Forward Loss Weight & 0.2 \\
        Learning Rate & $10^{-4}$ \\ 
    \bottomrule
    \end{tabular}
    \caption{ICM hyperparameters}
    \label{tab:icm-param}
\end{table}

\subsection{Training Dynamics}
Table~\ref{tab:pre-n-train-hyper} shows the hyperparameters for the training routine:
\begin{table}[htbp]
    \centering
    \begin{tabular}{cc}
    \toprule
        Name &  Value  \\
    \midrule
        Epoch Length     & 28800\\
        Train Frequency  & 2400 \\
        Maximum Eposide length & 1200 \\
        Parallel env thread count & 8 \\
        Batch size       & 128 \\
        Batches per Train & 20 \\
        Episodes per test & 100 \\
    \bottomrule
    \end{tabular}
    \caption{Training routing hyperparameters}
    \label{tab:pre-n-train-hyper}
\end{table}

We use different epoch lengths and maximum episode length for the post-novelty situation: \newpage
\begin{table}[htbp]
    \centering
    \begin{tabular}{cc}
    \toprule
        Name &  Value  \\
    \midrule
        Epoch Length     & 4800\\
        Train Frequency  & 800 \\
        Maximum Eposide length & 400 \\
        Parallel env thread count & 4 \\
        Batch size       & 128 \\
        Batches per Train & 8 \\
        Episodes per test & 100 \\
    \bottomrule
    \end{tabular}
    \caption{Post-novelty training routine hyperparameters}
    \label{tab:post-n-train-hyper}
\end{table}

\section{Novelty Implementation Example}
Here is an implementation of an example novelty, which makes the traders busy $50\%$ of the time (and they won't trade with you when they are busy.)

\subsection{Configuration file}
This is the content of the novelty configuration YAML file. It specifies that the novelty needs to be injected at episode 0, and updates the action module associated with the \texttt{trade} action to be the new \texttt{BusyTrade} module. It also supplies the parameter of the new trade module, the percentage of time when the traders are busy.
\begin{lstlisting}
---
novelties:
  '0':
    actions:
      trade:
        module: novelties.evaluation1.busy_traders.trade_busy.BusyTrade
        busy_ratio: 0.5
\end{lstlisting}

\subsection{\texttt{trade.py} action module file}
This is the content of the modified \texttt{Trade} action module, which inherits from the default module. The new module raises an error when the internally generated random number is below the $50\%$ threshold, and gives the error message saying that it is busy.
\begin{lstlisting}[language=Python]
from gym_novel_gridworlds2.actions.action import PreconditionNotMetError
from gym_novel_gridworlds2.contrib.polycraft.actions import Trade

import numpy as np

class BusyTrade(Trade):
    def __init__(self, busy_ratio=0, *args, **kwargs):
        self.busy_ratio = busy_ratio
        super().__init__(*args, **kwargs)

    def do_action(self, agent_entity, target_type=None, target_object=None, **kwargs):
        threashold = self.dynamics.rng.uniform(0, 1)
        if threashold < self.busy_ratio:
            raise PreconditionNotMetError("Trader is busy. Please try again later.")
        return super().do_action(agent_entity, target_type, target_object, **kwargs)

\end{lstlisting}

More examples of the novelty can be found in our repository.


\end{document}